\documentclass[letterpaper, 10 pt, conference]{ieeeconf}  

\IEEEoverridecommandlockouts                              

\usepackage{graphics} 
\usepackage{epsfig} 
\usepackage{mathptmx} 
\usepackage{times} 
\usepackage{float}
\usepackage{bbding}
\usepackage{amsmath} 
\usepackage{ragged2e} 
\usepackage{amssymb}  
\usepackage{multirow}
\usepackage{booktabs}
\usepackage{makecell}
\usepackage{caption}
\usepackage{subfigure}
\usepackage{multicol,float}
\usepackage{cite}

\title{\LARGE \bf
Gated Driver Attention Predictor
}

\author{Tianci Zhao$^{1}$, Xue Bai$^{2}$, Jianwu Fang$^{1}$ and Jianru Xue$^{3}$
\thanks{$^{1}$T. Zhao and J. Fang are with the College of Transportation Engineering, Chang'an University, Xi'an, China
        {\tt\small fangjianwu@chd.edu.cn.}}%
        \thanks{$^{2}$X. Bai is with the Science and Technology on Complex System Control and Intelligent Agent Cooperative Laboratory, Beijing, China
        {\tt\small 18829281638@163.com.}}%
\thanks{$^{3}$J. Xue is with the College of Artificial Intelligence, Xi'an Jiaotong University, Xi'an, China
        {\tt\small jrxue@mail.xjtu.edu.cn.}}%
        }

\begin{document}

\maketitle
\thispagestyle{empty}
\pagestyle{empty}

\begin{abstract}
Driver attention prediction implies the intention understanding of where the driver intends to go and what object the driver concerned about, which commonly provides a driving task-guided traffic scene understanding. Some recent works explore driver attention prediction in critical or accident scenarios and find a positive role in helping accident prediction, while the promotion ability is constrained by the prediction accuracy of driver attention maps. In this work, we explore the network connection gating mechanism for driver attention prediction (Gate-DAP). Gate-DAP aims to learn the importance of different spatial, temporal, and modality information in driving scenarios with various road types, occasions, and light and weather conditions. The network connection gating in Gate-DAP consists of a spatial encoding network gating, long-short-term memory network gating, and information type gating modules. Each connection gating operation is plug-and-play and can be flexibly assembled, which makes the architecture of Gate-DAP transparent for evaluating different spatial, temporal, and information types for driver attention prediction. Evaluations on DADA-2000 and BDDA datasets verify the superiority of the proposed method with the comparison with state-of-the-art approaches.
\end{abstract}



\section{Introduction}
\label{section1}
The interaction between the driver and the surrounding road environment implies frequent intention prediction. The driver fixation contains the intention of where to intend to go or be interested in safe decision-making. Driver attention is a typical cognition load that reflects the capacity for selecting and perceiving the useful road context \cite{DBLP:journals/tits/RasouliT20}, and is investigated largely for normal driving situations \cite{DBLP:journals/tits/TangYS22}. With an important expansion, recent researches find that driver attention shows manifested promotion for accident prediction in driving scenes \cite{DBLP:conf/iccv/Bao0K21,DBLP:conf/itsc/FangYQXWL19}. Driver attention prediction can help to find the crashing (to be involved in an accident) object in advance under many adverse environment conditions \cite{DBLP:journals/tits/FangYQXY22,gan2022multisource}.

The popular prototype in this topic is to leverage the powerful fitting ability of deep learning architectures. In different driving scenarios, different drivers may focus on different scene regions because of their subjective will, which makes the driver attention prediction with large prediction uncertainty. Consequently, some works begin to adopt multitudinous information, such as \emph{road semantics}, \emph{scene motion}, \emph{intended goals}, \emph{object locations},  \emph{etc.}, to weaken the prediction uncertainty and find the key elements in accident scenarios for driver attention prediction \cite{DBLP:conf/iccv/BaeePK0OB21}.

 \begin{figure}[!t]
  \centering
 \includegraphics[width=\hsize]{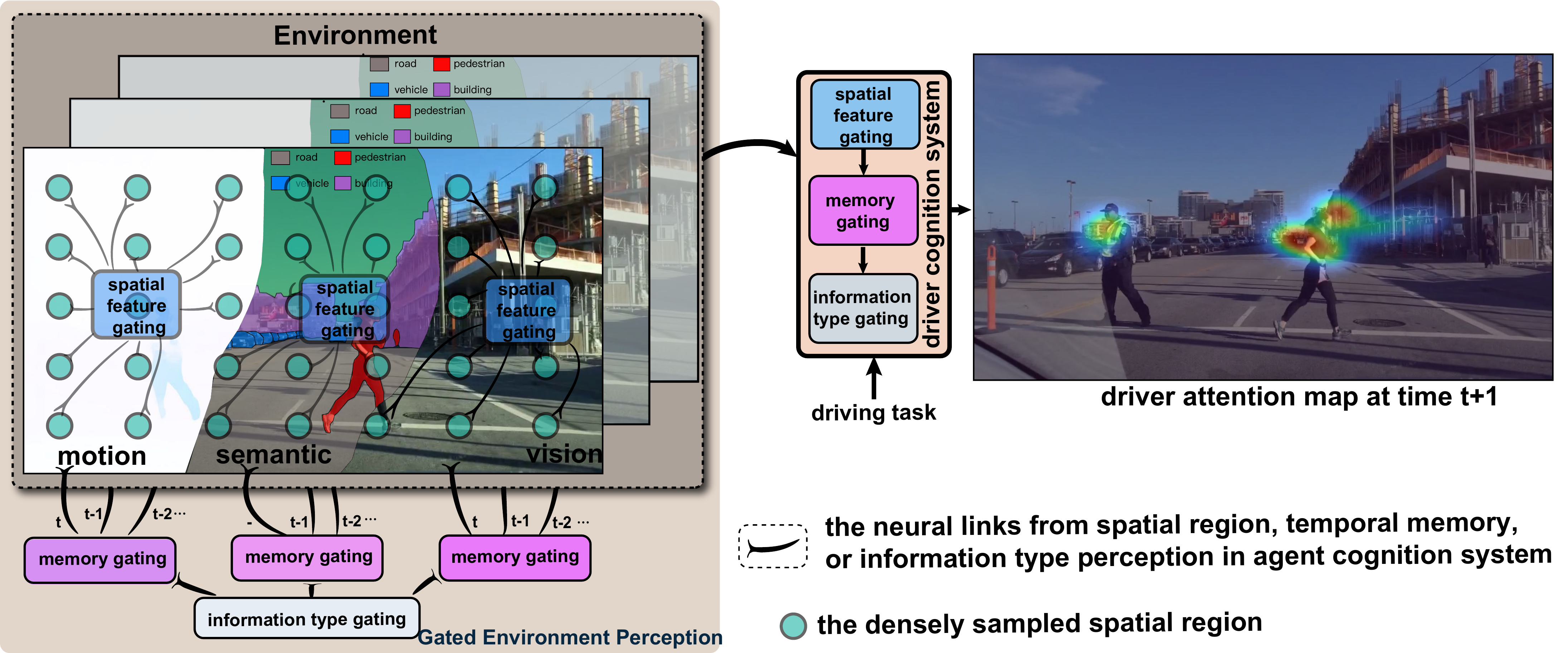}
  \caption{\small{Illustration with a critical scenario in BDDA dataset \cite{DBLP:conf/accv/XiaZKNZW18}, where we formulate a driver cognition system with the neural links for spatial feature, temporal memory, and information type (e.g., vision, semantics, and motion) selection (i.e., by gating functions).}}
  \label{fig1}
  \vspace{-1em}
\end{figure}
Although existing works improve the performance of driver attention prediction, most of them are not explainable for which kinds of information or what module play the key role in the improvement. Commonly, fusing all information or integrating all modules is universal for final implementation. In fact, it is natural that different scene information implies different promotion abilities from the aspects of the spatial region, temporal memory, or information types. Claimed by the recent research that appeared in Science \cite{Sci2022, Sci20222}, \emph{no neuron is an island}, and the outcomes of the exhibited human behaviors are driven by the connections between the neurons. The connection of the brain neurons stimulates human intelligence. How to leverage this finding, measure, or discover the expressive ability of different information and encoding modules in this topic? This question implies information selection or counterfactual reasoning problems \cite{DBLP:conf/eccv/JacobZBCPC22}. The work of \cite{DBLP:journals/tits/TangYS22} explores the driver's visual fixation behavior with a model-driven white-box representation, which introduces the driving task-aware representation, such as steering angles, speed, etc. Then, the driving task-driven representation is fed into a weight learning module with motion and bottom-up saliency maps to select the information. Some works explore the counterfactual role of different input information by masking or changing some types of them \cite{DBLP:conf/iros/0006CC20,DBLP:conf/eccv/JacobZBCPC22}. These formulations involve many ad-hoc enumeration tricks, which introduce further questions for different masking strategies.

In this work, we aim to explore the network connection gating mechanism, for finding the flexible and transparent architecture for driver attention prediction (called Gate-DAP). Specifically, we design the network connection gating from the Spatial Feature Gating (SpaG), Long-Short-Term Memory Gating (MemoG), and Information-Type Gating (InfoG) (to be described in Sec. \ref{ben}), as illustrated in Fig \ref{fig1}. The connection gating units in Gate-DAP can be flexibly assembled for checking the roles of different information and encoding modules. In this work, we introduce four types of information of RGB video frames, road semantic images, optical flow (motion) images, and the drivable road area. Different types of information are encoded with the Vision Transformer (ViT) to leverage the self-attention mechanism. To conveniently evaluate the role of different information, we introduce an object-centric counterfactual analysis to check the role of certain types of information but maintain the whole model unchanged. The \textbf{contributions} are threefold.

\begin{itemize}
\item We explore the network connection gating mechanism for achieving a transparent architecture of driver attention prediction (Gate-DAP)\footnote{The code will be available in \underline{https://github.com/JWFangit/Gate-DAP}.}, which refers to the spatial feature encoding network, temporal memory encoding network, and information fusion network.  
\item The connection gating units are plug-and-play and can be flexibly assembled. We introduce the object-centric counterfactual analysis for the information importance evaluation, which avoids to re-train the whole model with different configurations.
\item We evaluate the performance on two datasets, DADA-2000 \cite{DBLP:journals/tits/FangYQXY22} and BDDA \cite{DBLP:conf/accv/XiaZKNZW18}, and superior performance to other state-of-the-art methods is obtained. 
\end{itemize}

\section{Precedent Work}
\label{relatework} 
\subsection{Driver Attention Prediction in Driving Scenes}
The prediction of driver attention reflects the intention prediction of where intends to go or what is of interest in driving scenes and is investigated in many applications,  such as important object detection \cite{DBLP:journals/pacmhci/RongKFK22}, driver distraction detection \cite{DBLP:journals/tits/HuangF22}, driving model, or policy learning \cite{DBLP:journals/tits/MorandoVD19}. With the emergence of some large-scale benchmarks, such as the ones for normal driving situations (DR(eye)VE \cite{DBLP:journals/pami/PalazziACSC19}, TrafficGaze \cite{DBLP:journals/tits/DengYQNM20}, and CoCAtt \cite{DBLP:journals/corr/abs-2111-10014}) and the critical or accident scenarios (Eyecar \cite{DBLP:conf/iccv/BaeePK0OB21}, DADA-2000 \cite{DBLP:journals/tits/FangYQXY22}, and BDDA \cite{DBLP:conf/accv/XiaZKNZW18}), the driver attention prediction has fast progress in recent years. Driver attention prediction in driving scenes exhibits three kinds of research prototypes: \emph{data-driven}, \emph{model-driven}, and \emph{cognitive-conditioned} approaches. 

\textbf{Data-driven formulation} aims to leverage the data distribution of driver attention in different scenarios or datasets. The motion, semantic, and RGB frames are commonly used in the data encoding modules \cite{DBLP:conf/iccv/BaeePK0OB21,DBLP:journals/pami/PalazziACSC19}. For example, Deng \emph{et al.} \cite{DBLP:journals/tits/DengYQNM20,DBLP:journals/ieeejas/TianDY22} collect the driver attention data the targeting highway scenario and the rainy condition, and some Convolution Neural Network (CNN) models are proposed for the video frame encoder and driver attention map decoder. 

\textbf{Model-driven formulation} is accompanied by the innovation of many kinds of learning models for learning the driver attention patterns, such as the conditional Generative Adversarial Network (GAN), Inverse Reinforcement Learning (IRL) \cite{DBLP:conf/iccv/BaeePK0OB21}, or some explainable models \cite{DBLP:journals/tits/TangYS22}, etc.

\textbf{Cognitive-conditioned approaches} explore the driver status in driver attention prediction, such as the distraction state and the intention, to guide the drive attention prediction. CoCAtt \cite{DBLP:journals/corr/abs-2111-10014} investigates the driver attention pattern in turning behavior or straight-moving intention. Huang and Fu \cite{DBLP:journals/tits/HuangF22} adopt the predicted driver attention map to detect the state of driver distraction. Analogously, the driver's attention is also estimated by the head pose \cite{DBLP:journals/corr/abs-2012-12754}.

The aforementioned works for driver attention prediction show manifested progress, while most of them are not explainable for which kinds of information or what module plays the key role in the improvement.

\subsection{Gating Networks}
In recent years, many kinds of Dynamic Neural Networks (DNN) \cite{DBLP:journals/pami/HanHSYWW22} have been proposed for fusing different information with various fusion strategies. Gating networks are the types of typical paradigms. Different from the previous attention-based works \cite{DBLP:conf/aaai/WangZHZS20}, the gating network adopts different gating functions to restrain the link of encoding networks in spatial, temporal, or modality aspects. Gating networks usually concentrate on the feature layer skipping  \cite{DBLP:conf/eccv/WangYDDG18}, feature channel gating \cite{DBLP:conf/cvpr/LiWWLLC21}, network path selection \cite{DBLP:conf/cvpr/LiSCLZWS20}, and information type allocation \cite{DBLP:conf/iclr/MengPLSKSOF21}. Commonly, the gating functions are added as a lateral skip residual link on the original feature extraction pathways. The gating function is a plug-in module that can be used in arbitrary locations in different networks.

\textbf{Spatial gating networks} can be divided into pixel-level gating networks, region-level gating networks, and scale-level gating networks. For example, gated convolution \cite{DBLP:conf/iccv/YuLYSLH19} is one typical pixel-level gating function by learning a soft mask from the data, which achieves a dynamic spatial feature selection for each feature channel and spatial location.

\textbf{Temporal gating networks} commonly add the gating function in the hidden state or the input of Recurrent Neural Networks (RNN). For example, Wu \emph{et al.} propose an efficient video recognition method, which uses a conditional gating module to decide whether more discriminative information is needed for the current video frame.

\textbf{Modality gating networks} aim to explore the modality selection for final decisions. For example, Dynamic Multimodal Fusion (DynMM) \cite{DBLP:journals/corr/abs-2204-00102} proposes a text-vision-audio fusion method for the final decision, which has a gating network for selecting the expert networks on each modality.

Most gating networks explore spatial, temporal, and modality gating separately, while different information is woven together and each kind of gating function may have an influence on each other. In this work, we explore the spatial, temporal, and information types gating together for verifying different encoding modules for driver attention prediction.
  \begin{figure*}[!t]
  \centering
 \includegraphics[width=\hsize]{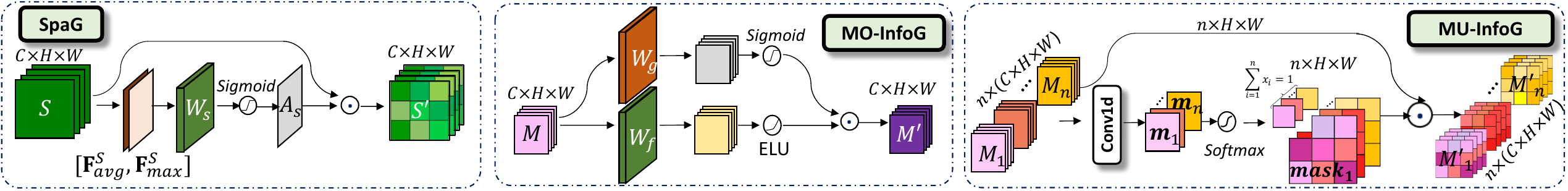}
  \caption{\small{The structure of SpaG, MO-InfoG, MU-InfoG, where $\odot$ is the Hadamard product.}}
  \label{fig2}
\end{figure*}

\section{Gate-DAP}
\label{method}
In this section, we first describe the network connection gating functions, and then present the whole model of driver attention prediction. The network connection gating is inspired by the connection mechanism between the neurons in the brain, where each kind of connection stimulates the intelligent understanding of different information. Meanwhile, the gating fulfills an information selection. 

\subsection{Network Connection Gating Functions}
\label{ben}
This work leverages the gating mechanism from the aspects of spatial region selection, temporal memory selection, and information type selection, and fulfills them by the Spatial Region Gating (\textbf{SpaG}), Long-Short-Term Memory Gating (\textbf{MemoG}), and Information-Type Gating (\textbf{InfoG}) (as shown in Fig. \ref{fig2}) to be described in following.

\subsubsection{SpaG}
\label{3.3}
SpaG aims to select the spatial region feature for subsequent information encoding and driver attention prediction. SpaG is fulfilled by a spatial attention module, which multiplies the generated feature attention tensors with the original feature tensor to achieve spatial gating of each image patch.
As shown by the SpaG structure in Fig. \ref{fig2}, for the input feature tensor ${\bf{S}}\in\mathbb{R}^{{C}\times{W}\times{H}}$, where ${C}$, ${H}$, and ${W}$ are the number of feature channels, the height, and width of the feature map, respectively. Similar to \cite{DBLP:conf/eccv/WooPLK18}, we first use the global average-pooling (GAP) and global max-pooling (GMP) operations along the channel axes for highlighting the spatial activation regions, and generate two types of 2D feature maps, i.e., ${\bf{F}}_{avg}^{S}\in\mathbb{R}^{{1}\times{W}\times{H}}$ and ${\bf{F}}_{max}^{S}\in\mathbb{R}^{{1}\times{W}\times{H}}$. Based on the spatial attention convolution, these 2D feature maps are then concatenated and convolved by a standard convolution layer, producing a 2D spatial attention map ${A}_{s}=\sigma({W}_s([{{\bf{F}}_{avg}^{S};{\bf{F}}_{max}^{S}}])) \in\mathbb{R}^{{W}\times{H}}$, where [;] is the concatenation, $\sigma$ is the \emph{Sigmoid} function and ${W}_s$ is the weight of a convolution operation. Then the final output ${\bf{S}}$ after spatial gated convolution is:
\begin{equation}
{\bf{S}}'={A}_{s}\odot{\bf{S}}.
\label{eq:4}
\end{equation}

\subsubsection{InfoG}
\label{3.1}
InfoG concentrates on the selection of different types of information, such as RGB video frames, motion features, semantic features, etc. It is inspired by that different types of information in the same driving scene may have differing importance for drivers. For this purpose, we design two kinds of information-type gating functions: Multiple Information Type Gating (MU-InfoG) and Monocular Information Type Gating (MO-InfoG). MU-InfoG fulfills a cross-attention model among different types of information, and MO-InfoG gates the single type of information.

\textbf{MO-InfoG:} MO-InfoG filters the input feature tensor ${\bf{M}}\in\mathbb{R}^{{C}\times{W}\times{H}}$ by:
\begin{gather}
{\bf{M}}'=\phi({W}_{f}\cdot{{\bf{M}}})\odot\sigma({W}_{g}\cdot{{\bf{M}}}),
\label{eq:1}
\end{gather}
where ${\bf{M}}'$ is the output after MO-InfoG, $\phi$ can be any activation function (e.g., ReLU, LeakyReLU, etc.), and the ELU activation function \cite{DBLP:journals/corr/ClevertUH15} is chosen in this work for relaxing the gradient and making the neuron be active all the time. ELU activation function is defined as $e^x-1$ for $x<0$ and $x$ for $x>=0$. ${W}_{g}$ and ${W}_{f}$ are the gated convolution filters \cite{DBLP:conf/iccv/YuLYSLH19} and the original feature convolution filters, respectively. Here, $\sigma(.)$ restrains the output of the gating value to $[0,1]$.

\textbf{MU-InfoG}: As shown in Fig. \ref{fig2}, for the input feature tensor ${\bf{M}}_{i} \in\mathbb{R}^{{C}\times{W}\times{H}}$, we first use 1D convolution to reduce their channel dimension and generate ${\bf{m}}_i \in \mathbb{R}^{{1}\times{W}\times{H}}$. Then, we concatenate ${\bf{m}}_i$ along the information type dimension to obtain a tensor ${\bf{MU}}$ with the size of $n\times W \times H$, where $n$ is the number of information types. Next, we use the \emph{softmax} function to ensure that the sum of the feature values of ${\bf{MU}}$ at each spatial dimension is 1 to achieve the feature selection. Finally, we rearrange ${\bf{MU}}$ to the shape of the original feature tensor ${\bf{M}}_{i}$ and obtained $n$ masks $\{{\bf{mask}}_i\}_{i=1}^n$ with size of ${H \times W}$. The output of MU-InfoG for each kind of information is denoted as ${\bf{M}}_i'$.

\subsubsection{MemoG}
\label{3.2}
  \begin{figure}[!t]
  \centering
 \includegraphics[width=\hsize]{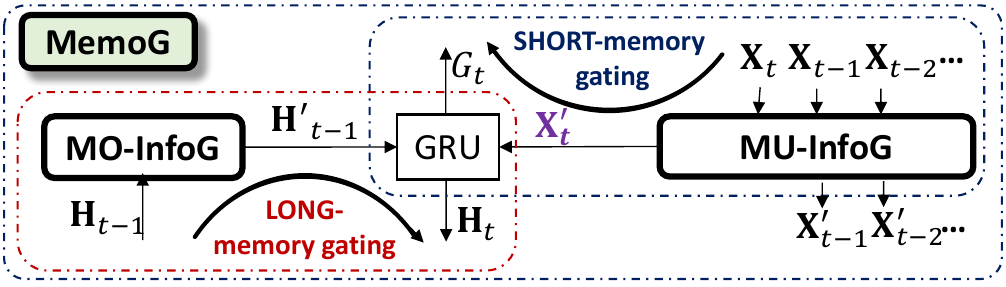}
  \caption{\small{The structure of MemoG, where $G_t$ is the output label of ${\bf{X}}_t'$ and omitted in this work.}}
  \label{fig3}
  \vspace{-1em}
\end{figure}
MemoG focuses on the temporal memory gating with a long and short window consideration. MemoG stands at the gate recurrent unit (GRU) and filters out redundant features in historical video frames with short and long-term information gating. We treat the hidden state ${\bf{H}}_t$ as a representation of a long-term memory at time $t$ after several times of temporal recurrences. The input ${\bf{X}}_t$ at time $t$ is denoted as the representation of short-term memory. 

Specially, in short-term memory, we consider the \textbf{uncertainty} in driver attention prediction, which is fulfilled by the cross-attention of the input feature tensors of $k$ frames, i.e., $[{\bf{X}}_t,{\bf{X}}_{t-1},...,{\bf{X}}_{t-k}]$. If we encounter a sudden change in driving scenes, the uncertainty estimation will give a large weight to the frame with sudden change. This consideration is achieved by the MU-InfoG operation on $[{\bf{X}}_t,{\bf{X}}_{t-1},...,{\bf{X}}_{t-k}]$ and generates $[{\bf{X}}_t',{\bf{X}}_{t-1}',...,{\bf{X}}_{t-k}']$.

For long-term memory, we employ the MO-InfoG to gate the hidden state ${\bf{H}}_{t-1}$ obtained in the previous time. Consequently, the MemoG is modeled by:
\begin{gather}
{\bf{H}'}_{t-1}={\text{MO-InfoG}}({\bf{H}}_{t-1}), {\bf{H}}_{t}=\text{GRU}({\bf{H}'}_{t-1},{\bf{X}'}_{t}),
\label{eq:3}
\end{gather}
Fig. \ref{fig3} demonstrates the structure of MemoG.
We omit $Y_t$ for the driver attention decoding. We explicitly enforce the gating function to the input hidden state and current observation, which aims to purify the temporal information before the GRU unit (with 256 dimensions of hidden state).

Thus, the SpaG, InfoG, and MemoG are described, which can be flexibly adopted in any deep learning model, and are carefully utilized in our driver attention prediction network.

\subsection{Driver Attention Prediction}

\begin{figure}[!t]
  \centering
 \includegraphics[width=\hsize]{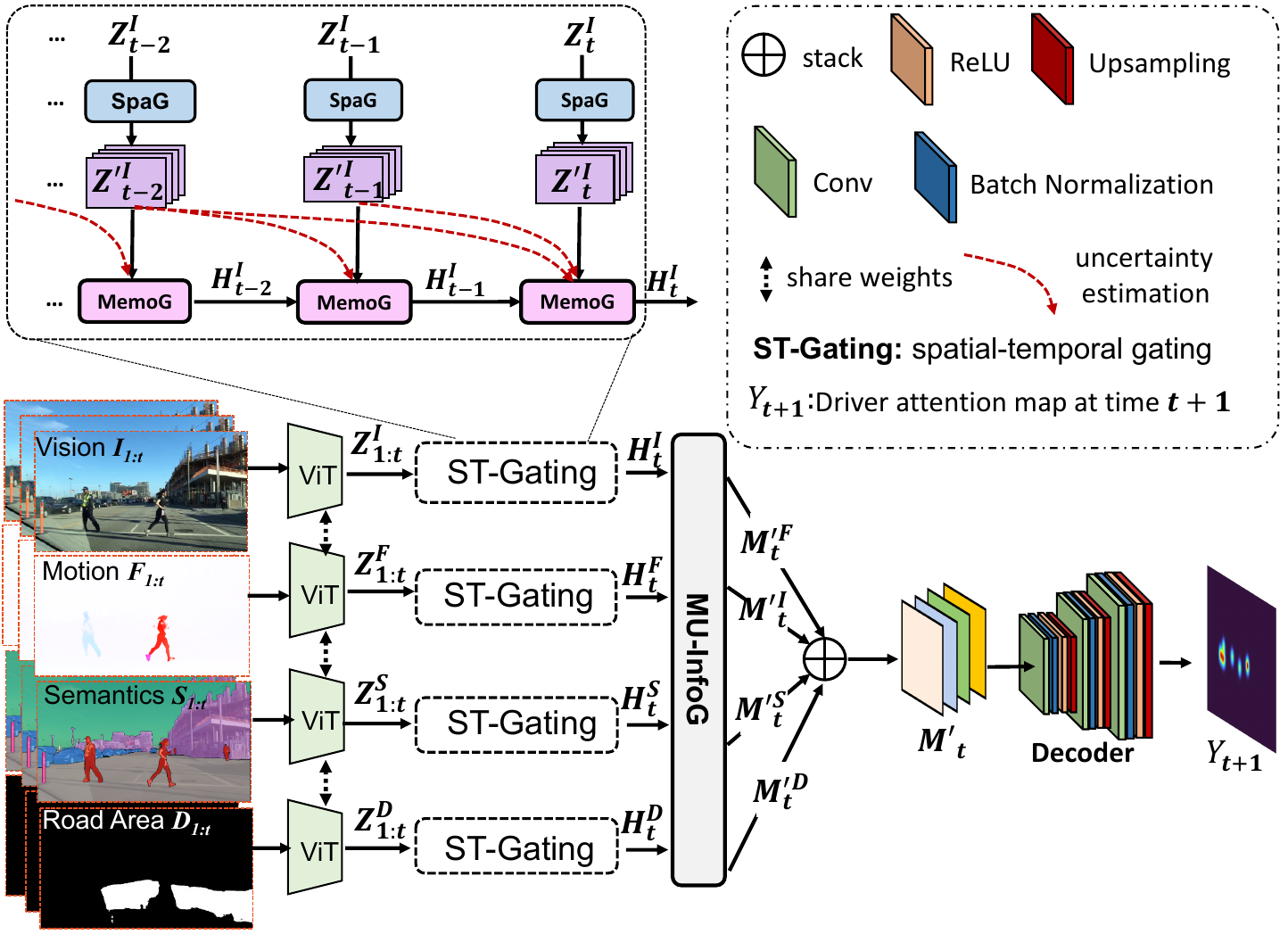}
  \caption{The pipeline of \textbf{Gate-DAP}. Notably, we estimate the uncertainty in the memory gating module, which is useful for feature learning with sudden scene changes.}
  \label{fig4}
    \vspace{-1em}
\end{figure}

In driver attention prediction tasks for driving scenarios, the driver's eye movements are usually influenced by multiple factors due to complex and variable road conditions. The whole pipeline of the Gated Driver Attention Predictor (Gate-DAP) model is shown in Fig. \ref{fig4}. In this work, we consider four kinds of information to model the driving scenes, and each input sample clip of Gate-DAP consists of a group of [${I_{1:t}}$, ${F_{1:t}}$, ${S_{1:t}}$, ${D_{1:t}}$], where the $t^{th}$ frame at one clip is denoted as ${I_t}$ for RGB information, ${F_t}$ for motion information, ${S_t}$ for semantic information, and ${D_t}$ for drivable region information. Notably, motion frame ${F_t}$ is obtained by computing the motion correlation between ${I_t}$ and ${I_{t-1}}$. 

\textbf{Backbone Model:} Each frame in one clip is encoded by a backbone model, respectively. As shown in Fig. \ref{fig4}, different types of information follow a parallel encoding but share the encoder weights. Because the encoding of each type of information is the same, we take the $t^{th}$ RGB frame in one clip as an example. In this work, we take the vision-transformer model (ViT) as the backbone model, which is pre-trained in ImageNet-1K by masked auto-encoder (MAE) \cite{he2022masked}. ViT is fulfilled by a multi-head self-attention module on image patches with position embedding. In this work, the number of self-attentive heads of ViT is set as 12, the depth of layers is set as 12, and the patch size is 16.

\textbf{Connection Gating:} Denote the feature embedding of the RGB frame clip is $[{\bf{Z}}_1^I,...,{\bf{Z}}_t^I]$. The feature embedding of each frame is separately gated by the SpaG to achieve a spatial feature selection. The output of the SpaG at each frame is correlated by the MemoG to fulfill a temporal memory gating over the frames within the frame clip. MU-InfoG is adopted by following the MemoG at the $t^{th}$ frame for gating different types of information.  The gating part of Gate-DAP is:
\begin{gather}
{\bf{Z}}_{t}^{I}=\text{ViT}|_{\text{MAE}}(I_t),\\
{\bf{Z}'}_{t}^{I}=\text{SpaG}({\bf{Z}}_{t}^{I}),\\
{\bf{H}}_{t}^{I}=\text{MemoG}({\bf{H}}_{t-1}^{I}, [{\bf{Z}'}_{1}^{I},...,{\bf{Z}'}_{t}^{I}]),\\
[{\bf{M}'}^I_t,{\bf{M}'}^F_t,{\bf{M}'}^S_t,{\bf{M}'}^D_t]=\text{MU-InfoG}({\bf{H}}_{T}^{I},{\bf{H}}_{T}^{S}, {\bf{H}}_{T}^{F}, {\bf{H}}_{T}^{D}),\\
{\bf{M}'}_t=\text{Stack}[{\bf{M}'}^I_t,{\bf{M}'}^F_t,{\bf{M}'}^S_t,{\bf{M}'}^D_t].
\label{eq:1}
\end{gather}
where ${\bf{M}'}_t$ is the final feature representation at time $t$ after stacking all information types and adopted to decode the future attention map at time $t+1$. Because embedding each type of information is time-consuming by the ViT model, we share the weight for different types of information.

\textbf{Attention Map Decoding:} After the MU-InfoG operation, ${\bf{M}}'_t$ needs to be converted to an attention map with the same size as the input frame. Therefore, ${\bf{M}}^{'}_t$ is decoded as the final driver attention map $Y_{t+1}$ for the ${(t+1)^{th}}$ frame.

The decoding module consists of three interleaved blocks, each one of which contains a 2D convolution, batch normalization, ReLU function, and upsampling operation, fulfilled by [${conv(3\times3, 128)}$ $\rightarrow$ \emph{Batch Normalization} $\rightarrow$ ${ReLU}$ $\rightarrow$ ${upsampling]\times4}$ $\rightarrow$ ${conv(3\times3, 1)}$, and the final block ends with ${Sigmoid}$ function for mapping the output value to [0,1] to highlight the focused regions of the ${(t+1)^{th}}$ frame.

\textbf{Loss Function:} 
Similar to DADA \cite{DBLP:journals/tits/FangYQXY22}, we also take the joint loss function to train the gated network, which contains the Kullback-Leibler distance (${KLD}$), linear correlation coefficient (${CC}$), and normalized scan path significance (${NSS}$). The specific loss function is denoted as:
\begin{equation}\small
\begin{array}{ll}
\mathcal{L}({Y}_{t+1},\hat{Y}_{t+1})=\sum_{i=1}^{N}{Y_{t+1}}({i}){log}(\varepsilon+\frac{Y_{t+1}(i)}{\varepsilon+\hat{Y}_{t+1}({i})})-\alpha\cdot\frac{{cov}({Y_{t+1}},\hat{Y}_{t+1})}{{\rho(Y_{t+1})}{\rho}(\hat{Y}_{t+1})}\\ \verb'    '-\beta\cdot\frac{1}{\sum_{i=1}{P_{t+1}(i)}}\sum_{i=1}\frac{\hat{Y}_{t+1}{(i)}-\mu(\hat{Y}_{t+1})}{{\rho}(\hat{Y})}{P_{t+1}(i)}
\label{eq:7}
\end{array}
\end{equation}
where $Y_{t+1}$ and $P_{t+1}$ are ground-truth saliency and fixation point maps, respectively, $\hat{Y}_{t+1}$ is the predicted attention map. $i$ indexes the $i^{th}$ pixel across the all $N$ pixels of the saliency map. $\alpha$ and $\beta$  are the coefficient that adjusts the weight of ${CC}$ and ${NSS}$, ${N}$ indicates the number of image pixel points, ${cov}({Y},\hat{Y})$ indicates the covariance of ${Y}_{t+1}$ and $\hat{Y}_{t+1}$; $\rho$ indicates standard deviation operation, and $\varepsilon$ is a very small constant to prevent the operation from errors such as the number of denominators being 0.

\section{Experiments}
\label{expe}
\subsection{Dataset}

In this paper, we evaluate the performance of the proposed Gate-DAP on two challenging datasets with critical or accident scenarios, i.e., BDD-A \cite{DBLP:conf/accv/XiaZKNZW18} and DADA-2000 \cite{DBLP:journals/tits/FangYQXY22}. 

BDD-A \cite{DBLP:conf/accv/XiaZKNZW18} consists of 1,232 sequences (each one owns about 10 seconds). It focuses on critical situations such as occlusions, truncations, and emergency braking. To obtain annotations, 45 drivers are asked to watch videos, and their eye movements are recorded by an eye tracker to generate fixations. We follow its partition and obtain 28k frames for training, 6k frames for validation, and 9k frames for testing. 

DADA-2000 \cite{DBLP:journals/tits/FangYQXY22} concentrates on driver attention prediction in accident scenarios. This dataset contains 2000 videos with over 658,746 frames. We follow the work \cite{DBLP:journals/tits/FangYQXY22} that 1000 videos are used for performance evaluation, which provides 598 training sequences (about 214k frames) and 222 testing sequences (about 70k frames), respectively. 

\subsection{Implementation Details}

The proposed method is implemented using the PyTorch framework. During the training process, we used the Adam optimizer with a learning rate of $10^{-6}$ and a weight decay of 0.0001. The entire model is trained in end-to-end mode, and the entire training process takes about 20 hours and 6 hours on one NVIDIA RTX2080Ti GPU with 11GB RAMs for DADA and BDD-A datasets, respectively. In addition, regarding the number of input frames in one clip, based on our previous research \cite{DBLP:journals/tits/FangYQXY22,DBLP:journals/tcsv/LiFXX21}, we found that for the frame or map prediction problems, more input frames will consume more computing resources with little performance gain. Therefore, in our implementation, due to the limitation of RAM space, each input clip contains 4 consecutive frames. We pre-prepare the semantic images, optical flow images, and drivable area images in advance using the DeeplabV3 \cite{DBLP:journals/corr/ChenPSA17},  FlowNet2.0 \cite{ilg2017flownet}, and Yolo-P \cite{wu2021yolop}, respectively.

\textbf{Metrics:} Following the previous driver attention prediction methods \cite{DBLP:journals/tits/FangYQXY22,DBLP:conf/iccv/BaeePK0OB21,DBLP:journals/pami/PalazziACSC19}, we utilize five metrics to evaluate the performance, which contains three distribution-based metrics, i.e., Kullback-Leibler Divergence (KLD), Pearson Correlation Coefficient (CC), and Similarity (SIM), and two location-based metrics, i.e., Normalized Scanpath Saliency (NSS) and the area under the receiver operating characteristic (ROC) curve (AUC). Here, two variants of AUC were used, namely AUC-Judd (AUC-J) and shuffled AUC (AUC-S). 

\subsection{Ablation Study}
\begin{figure}[!t]
  \centering
 \includegraphics[width=\hsize]{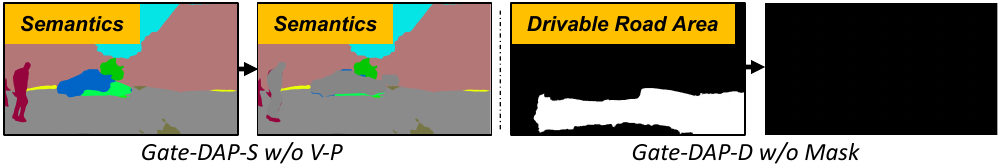}
  \caption{\small{The counterfactual operation on a semantic image (removing pedestrians and vehicles, i.e., ``\emph{Gate-DAP-S w/o V-P}") and removing the binary mask in drivable area image (i.e, ``\emph{Gate-DAP-D w/o Mask}").}}
  \label{fig5}
\end{figure}

\textbf{1) Which information is important? A counterfactual analysis.}
\label{counter}
To evaluate the importance of each type of information, this work introduces a counterfactual analysis strategy. We all know that most participants in driving scenes are pedestrians and vehicles, and these two types of semantics basically attract driver attention in most situations. If we remove these two kinds of semantics in the images, the input images may only contain the background. Specifically, we maintain the whole architecture of Gate-DAP and remove these two kinds of semantics one by one for each type of information (See Fig. \ref{fig5} for semantic information). For the drivable area image, we remove the binary mask. 

This strategy does not need to re-train the model with different information configurations and check the importance of each type of information by the metric value difference with Gate-DAP-Full-Model. Totally, we obtain ten versions after counterfactual analysis, denoted as three RGB versions (``\emph{Gate-DAP-I w/o P}", ``\emph{Gate-DAP-I w/o V}", and ``\emph{Gate-DAP-I w/o V-P}"), three motion versions (``\emph{Gate-DAP-F w/o P}", ``\emph{Gate-DAP-F w/o V}", and ``\emph{Gate-DAP-F w/o V-P}"), three semantic versions (``\emph{Gate-DAP-S w/o P}", ``\emph{Gate-DAP-S w/o V}", and ``\emph{Gate-DAP-S w/o V-P}"), and one drivable mask version (``\emph{Gate-DAP-D w/o Mask}"). Here, ``\emph{P}", ``\emph{V}", and ``\emph{Mask}" denote the indication of pedestrians, vehicles, and road mask regions, respectively. Larger differences mean that the information with our counterfactual operation has more importance. If one kind of information with counterfactual operation hardly affects the result of each metric, it is useless. 

  \begin{figure}[!t]
  \centering
 \includegraphics[width=\hsize]{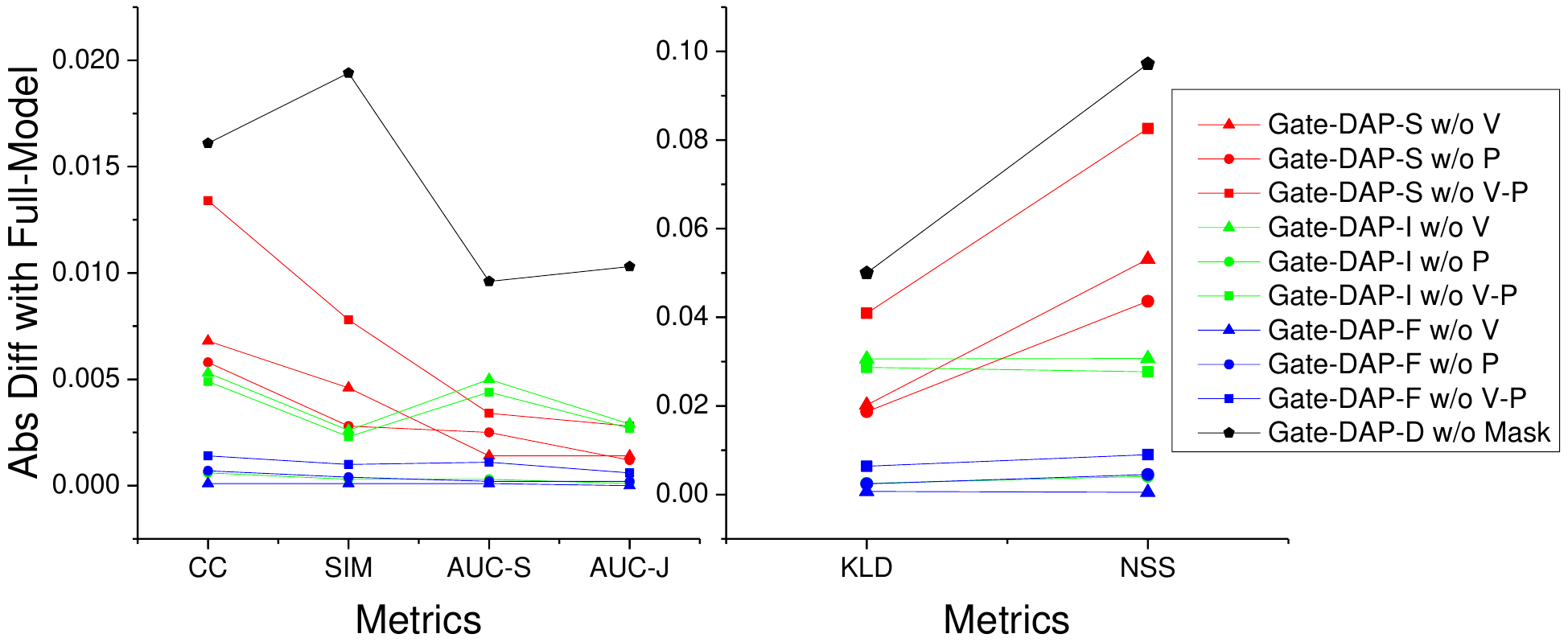}
  \caption{\small{Performance influence with the counterfactual operation on each kind of information, respectively. This evaluation is conducted on the testing set of the DADA-2000 dataset.}}
  \label{fig6}
\end{figure}

  \begin{figure*}[!t]
  \centering
 \includegraphics[width=\hsize]{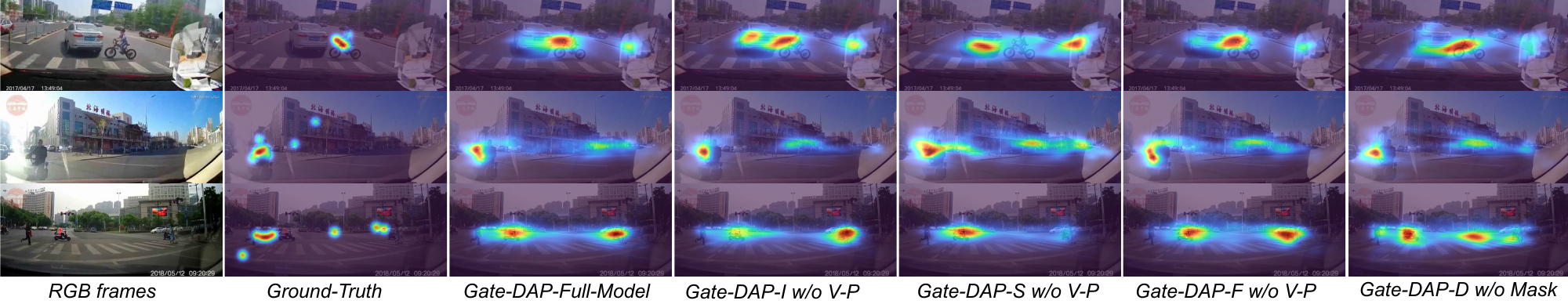}
  \caption{\small{Some predicted driver attention maps in DADA-2000 for checking the importance of different types of information.}}
  \label{fig7}
\end{figure*}

Fig. \ref{fig6} demonstrates the performance influence of each kind of information. We observe that the motion information in this work has the weakest influence on the performance, verified by ``\emph{Gate-DAP-F w/o P}",  ``\emph{Gate-DAP-F w/o V}", and ``\emph{Gate-DAP-F w/o V-P} with little difference. We also show some snapshots of predicted driver attention maps in Fig. \ref{fig7}. The visualization results demonstrate that removing the pedestrian and vehicles in the motion information (Fig. \ref{fig7}(f)) has little change with the Full-Model. On the contrary, the drivable area mask (we denote it as an indirect driving task representation) has the largest performance influence (marked by the black lines in Fig. \ref{fig6}). Besides motion information, other kinds of information have an impact on performance to a large extent. Therefore, we think the role of motion information in this work is very little.

\begin{figure}[!t]
\begin{minipage}[b]{0.55\linewidth}
\centering
\includegraphics[width=\linewidth]{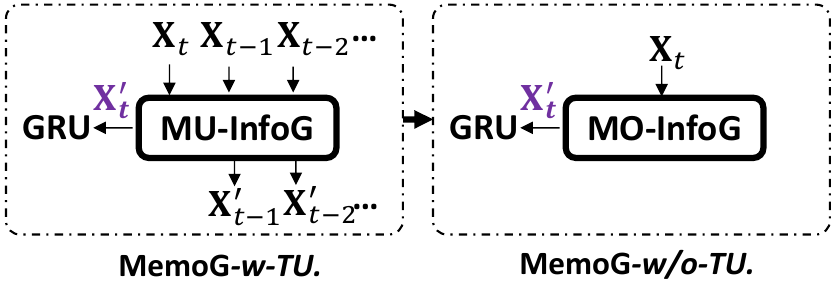}
\caption{\small{Difference of MemoG\emph{-w-TU.} and MemoG\emph{-w/o-TU.}}}
\vspace{-1.5em}
\label{fig8}
\end{minipage}
\begin{minipage}[b]{0.38\linewidth}
\centering
\scriptsize
 \renewcommand{\arraystretch}{1.2}
 \setlength{\tabcolsep}{0.3mm}{
 \centering
  \captionof{table}{\footnotesize{Performance evaluation of MemoG\emph{-w-TU.} and MemoG\emph{-w/o-TU. on DADA-2000.}}}
  \label{tab1}
\begin{tabular}{l|ccc}
\toprule[0.8pt]
 & $\text{KLD}\downarrow$ & $\text{CC}\uparrow$& $\text{NSS}\uparrow$ \\\hline
MemoG\emph{-w-TU.} & $1.66$ & $0.47$& $3.10$ \\\hline
MemoG\emph{-w/o-TU.}& $\textbf{1.65}$ & $\textbf{0.48}$ & $\textbf{3.14}$ \\ 
\toprule[0.8pt]
\end{tabular}}
\end{minipage}
\vspace{-1em}
\end{figure}

\textbf{2) Temporal uncertainty evaluation in MemoG.}
As aforementioned, we consider the temporal uncertainty in MemoG by cross-attention model for successive frames. This consideration aims to find the sudden scene change in critical or accident scenarios. Here, we evaluate the role of this consideration by the setting of MemoG with or without the Temporal Uncertainty (i.e., MemoG-w-TU. and MemoG-w/o-TU.). Accordingly, the structure of MemoG for the short-term memory gating is changed, as shown in Fig. \ref{fig8}.  Table. \ref{tab1} presents the results on the testing set of the DADA-2000 dataset, and we can see that \textbf{TU.} shows a slight promotion role for driver attention prediction.

\textbf{3) How about the role of different gating modules?} The primary insight of this work is to introduce the gating modules. To evaluate their roles, we take contrastive experiments, where we close the relative gating modules in the Gate-DAP model. Consequently, we have three versions in this evaluation, i.e., ``\emph{Gate-DAP w/o SpaG}", ``\emph{Gate-DAP w/o MemoG}", and ``\emph{Gate-DAP w/o MU-InfoG}". 

\emph{``Gate-DAP w/o SpaG"} is fulfilled by setting the weight map $W_s$ in Eq. 1 as an identity matrix.

\emph{``Gate-DAP w/o MemoG"} has two kinds of gating modules: MO-InfoG and MU-InfoG.  Eliminating MO-InfoG is achieved by setting the weight matrix $W_g$ and $W_f$ in Eq. 2 as two identity matrixes. Eliminating MU-InfoG is fulfilled by setting all the weight of different information equally. 

\emph{``Gate-DAP w/o MU-InfoG"} is obtained by the same setting for multiple kinds of information (with the same weight for each type of information).

\begin{figure}[!t]
  \centering
 \includegraphics[width=\hsize]{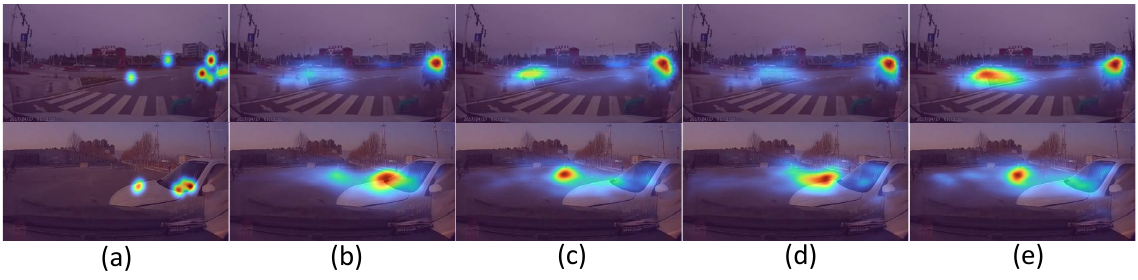}
  \caption{\small{The visualization of some predicted driver attention examples in DADA-2000 for evaluating different gating modules. (a): ground-truth; (b): Gate-DAP-Full-Model; (c): Gate-DAP w/o SpaG; (d): Gate-DAP w/o MemoG; (e): Gate-DAP w/o MU-InfoG.}}
  \label{fig9}
  \vspace{-1em}
  \end{figure}

The results are shown in Tab \ref{tab:2}, and we can see that the gating modules are positive for driver attention prediction. Among them, the MU-InfoG module has the greatest contribution. We can see that after adding the MU-InfoG module, all metric values improve. We also demonstrate some snapshots of predicted attention maps by different gating configurations in Fig \ref{fig9}. From the figure, we can clearly see without the SpaG and MU-InfoG, the predicted driver attention is intended on the road region, while the actual fixations concentrate on the vehicle. Gate-DAP with all gating modules can localize the true driver fixations well.
\begin{table}[!t]\scriptsize
\centering
\caption{\small{Evaluation on different gating modules.}}
\label{tab:2}
  \renewcommand{\arraystretch}{1.2}
 \setlength{\tabcolsep}{0.6mm}{
\begin{tabular}{ccc|cccc|ccc}
\toprule[0.8pt]
\multicolumn{3}{c|}{Gating Modules}& \multicolumn{4}{c|}{DADA2000} &  \multicolumn{3}{c}{BDD-A} \\ \hline
SpaG&MemoG&MU-InfoG& $\text{KLD}\downarrow$ & $\text{CC}\uparrow$ & $\text{SIM}\uparrow$ & $\text{NSS}\uparrow$ & $\text{KLD}\downarrow$ & $\text{CC}\uparrow$ & $\text{SIM}\uparrow$ \\ \hline
&&& $1.70$ & $0.47$ & $0.35$ & $3.07$& $1.47$& $0.52$& $0.40$ \\ 
\checkmark &&& $1.69$ & $0.47$ & $0.35$& $3.11$  & $1.44$& $0.53$& $0.42$\\ 
 &\checkmark&& $1.69$ & $0.47$ & $0.35$& $3.09$  & $1.46$& $0.52$& $0.40$\\
 && \checkmark& $1.67$ & ${0.48}$ & $0.35$& $3.13$  & $1.24$& $0.59$& $0.45$ \\
 \checkmark&\checkmark& & $1.68$ & $0.47$ & $0.35$& $3.13$ & $1.43$& $0.53$& $0.42$ \\
\checkmark&&\checkmark& $1.67$ & ${0.48}$ & $\textbf{0.36}$ & $3.13$& $1.19$& $0.59$& $0.47$ \\
&\checkmark&\checkmark& $1.67$ & ${0.48}$ & $\textbf{0.36}$& $3.12$ & $1.24$& $0.59$& $0.45$\\\hline
 \checkmark&\checkmark&\checkmark& $\textbf{1.65}$ & $\textbf{0.52}$ & $\textbf{0.36}$ & $\textbf{3.14}$ & $\textbf{1.12}$& $\textbf{0.61}$& $\textbf{0.49}$ \\ 
\toprule[0.8pt]
\end{tabular}}
\end{table}

\subsection{Comparison with State-of-The-Arts}
\label{Quanti}
To validate the superiority of Gate-DAP, seven representative driver attention prediction methods are compared, with five video-based BDDA \cite{DBLP:conf/accv/XiaZKNZW18}, DR (eye) VE \cite{DBLP:journals/pami/PalazziACSC19}, TwoStream \cite{zhang2018video}, SCAFNet \cite{DBLP:journals/tits/FangYQXY22}, TASEDNet \cite{min2019tased}, Vi-Net  \cite{jain2021vinet}, Flow-DA  \cite{sultana2023prediction}, and ASIAF-Net  \cite{DBLP:journals/tits/LiLCLLL22}. We compare the results reported in their works in this evaluation.

In addition, in the ablation studies, we introduce the counterfactual analysis and \emph{gating closing} to check the information's importance and gating roles, where different configurations are not re-trained. Certainly, to prove the reasonability of the ablation study ways, we re-train two versions, i.e., removing all gating modules (\emph{Gate-DAP w/o Gs}) and removing motion information (\emph{Gate-DAP w/o F}), to check whether our finding is reasonable or not.  The evaluation results are shown in Table. \ref{tab:3}. From the results, we can see that our Gate-DAP shows promising performance in all datasets, especially for the KLD metric and AUC-J metric. Besides, the results of \emph{Gate-DAP w/o Gs} and \emph{Gate-DAP w/o F} indicate that the ablation study ways in this work are promising for checking the role or importance of different model configurations without model re-training.

\begin{table}[!t]\scriptsize
\centering
\caption{\small{Comparison with several state-of-the-art methods.}}
\label{tab:3}
  \renewcommand{\arraystretch}{1.2}
 \setlength{\tabcolsep}{0.5mm}{
\begin{tabular}{c|cccccc|ccc}
\toprule[0.8pt]
 & \multicolumn{6}{c|}{DADA2000} &  \multicolumn{3}{c}{BDD-A} \\ \hline
 & $\text{KLD}\downarrow$ & $\text{CC}\uparrow$ & $\text{SIM}\uparrow$ & $\text{NSS}\uparrow$ & $\text{AUC-J}\uparrow$ & $\text{AUC-S}\uparrow$ & $\text{KLD}\downarrow$ & $\text{CC}\uparrow$ & $\text{SIM}\uparrow$ \\ \hline
DR(eye)VE \cite{DBLP:journals/pami/PalazziACSC19} & $2.27$ & $0.45$ & $0.32$ & $2.92$ & $0.91$ & $0.64$ & $1.95$ & $0.50$ & $-$\\ 
BDDA \cite{DBLP:conf/accv/XiaZKNZW18}& $3.32$ & $0.33$ & $0.25$ & $2.15$ & $0.86$ & $0.63$ & $1.49$ & $0.51$ & $0.35$\\
TwoStream \cite{zhang2018video} & $2.85$ & $0.23$ & $0.14$ & $1.48$ & $0.84$ & $0.64$ & $-$ & $-$ & $-$ \\
TASEDNet  \cite{min2019tased}& $1.78$ & $0.46$ & $0.31$ & $3.20$ & $0.92$ & $- $& $1.24$ & $0.55$ & $0.42$\\
Vi-Net  \cite{jain2021vinet}&  $-$& $-$ &  $-$& $-$  &  $-$ &  $-$& $1.39$ & 0.61 & 0.45\\ 
SCAFNet  \cite{DBLP:journals/tits/FangYQXY22}& $2.19$ & $0.50$ & $\textbf{0.37}$ & $3.34$ & $0.92$ & $0.66$ & $1.39$ & $0.54$ & $0.43$\\
Flow-DA  \cite{sultana2023prediction}&  $-$& $-$ &  $-$& $-$  &  $-$ &  $-$& $1.39$ & 0.61 & 0.45\\ 
ASIAF-Net  \cite{DBLP:journals/tits/LiLCLLL22}& $1.66$ & $0.49$ & $0.36$ & $\textbf{3.39}$ & $0.93$ & $0.78$ & $1.24$ & $\textbf{0.66}$ & $-$\\ \hline
Gate-DAP w/o Gs & $1.72$ & $0.46$ & $0.35$ & $3.03$ & $0.92$ & $0.84$ & $1.36$  &$0.54$  & $0.39$ \\
Gate-DAP w/o F& $1.66$ & $0.47$ & 0.35 & 3.12 & $0.92$ & $\textbf{0.85}$ & $1.18$ & $0.61$ & $0.42$\\
Gate-DAP & $\textbf{1.65}$ & $\textbf{0.52}$ & $0.36$ & $3.14$ & $\textbf{0.93}$ & $\textbf{0.85}$ & $\textbf{1.12}$ & $0.61$ & $\textbf{0.49}$\\
\toprule[0.8pt]
\end{tabular}}
\vspace{-2em}
\end{table}

\section{Conclusions}
\label{con}
This work proposes a Gated Driver Attention Predictor (Gate-DAP), which explores spatial feature gating, temporal memory gating, and information type gating to fulfill a transparent architecture of driver attention prediction networks. The gating modules are a plug-and-play that can be used to check the role of different kinds of features, i.e., spatial feature, temporal feature, and information type feature. In addition, we introduce a counterfactual analysis to evaluate the importance of different types of information. Based on the analysis, motion features have the least influence after removing the pedestrians and vehicles in motion images. On the contrary, drivable area images show manifest performance influence. Through the comparison with other state-of-the-art methods, Gate-DAP generates the best performance on DADA-2000 and BDDA datasets.

{\small
\bibliographystyle{IEEEtran}
\bibliography{ref}
}


\end{document}